\documentclass[10pt,twocolumn,letterpaper]{article}

\usepackage{iccv}
\usepackage{times}
\usepackage{epsfig}
\usepackage{graphicx}
\usepackage{amsmath}
\usepackage{amssymb}
\usepackage{url}
\usepackage{multirow}
\usepackage[noend]{algpseudocode}
\usepackage{algorithmicx,algorithm}
\usepackage{bm}
\usepackage{dsfont}
\usepackage{subfig}
\usepackage{epstopdf}

\newcommand{\tabincell}[2]{\begin{tabular}{@{}#1@{}}#2\end{tabular}}
\usepackage[breaklinks=true,bookmarks=false]{hyperref}

\iccvfinalcopy 

\def\iccvPaperID{4560} 

\ificcvfinal\fi
\begin{document}

\title{DUP-Net: Denoiser and Upsampler Network for 3D Adversarial Point Clouds Defense}

\author{Hang Zhou \quad Kejiang Chen \quad  Weiming Zhang\thanks{Corresponding author. } \quad Han Fang \quad Wenbo Zhou \quad  Nenghai Yu\\
University of Science and Technology of China\\
{\tt\small \{zh2991, chenkj, fanghan, welbeckz\}@mail.ustc.edu.cn \quad\quad \{zhangwm, ynh\}@ustc.edu.cn}
}

\maketitle
\ificcvfinal\thispagestyle{empty}\fi

\begin{abstract}
Neural networks are vulnerable to adversarial examples, which poses a threat to their application in security sensitive systems. We propose a Denoiser and UPsampler Network (DUP-Net) structure as defenses for 3D adversarial point cloud classification, where the two modules reconstruct surface smoothness by dropping or adding points. In this paper, statistical outlier removal (SOR) and a data-driven upsampling network are considered as denoiser and upsampler respectively. Compared with baseline defenses, DUP-Net has three advantages. First, with DUP-Net as a defense, the target model is more robust to white-box adversarial attacks. Second, the statistical outlier removal provides added robustness since it is a non-differentiable denoising operation. Third, the upsampler network can be trained on a small dataset and defends well against adversarial attacks generated from other point cloud datasets. We conduct various experiments to validate that DUP-Net is very effective as defense in practice. Our best defense eliminates 83.8\% of C\&W and $l_2$ loss based attack (point shifting), 50.0\% of C\&W and Hausdorff distance loss based attack (point adding) and 9.0\% of saliency map based attack (point dropping) under 200 dropped points on PointNet. 
\end{abstract}

\section{Introduction}
\emph{Deep Learning} has shown superior performance on several categories of machine learning problems, especially classification task. These \emph{Deep Neural Networks} (DNN) learn models from large training data to efficiently classify unseen samples with high accuracy. However, recent works have demonstrated that DNNs are vulnerable to \emph{adversarial examples}, which are maliciously created by adding imperceptible perturbations to the original input by attackers. Adversarially perturbed examples have been deployed to attack image classification service~\cite{liu2016delving}, speech recognition system~\cite{cisse2017houdini} and autonomous driving system~\cite{xiang2019generating}.

Heretofore, numerous algorithms have been proposed to generate adversarial examples for 2D images. When model parameters are known, a paradigm called \emph{white-box attacks} includes methods based on calculating the gradient of the network, such as FGSM ~\cite{goodfellow2014explaining}, IGSM ~\cite{gu2014towards} and JSMA ~\cite{papernot2016limitations}, and based on solving optimization problems, such as L-BFGS ~\cite{szegedy2013intriguing}, Deepfool ~\cite{moosavi2016deepfool} and Carlini \& Wagner (C\&W) attack ~\cite{carlini2017towards}. In the scenario where access to the model is not available, called \emph{black-box attacks}.

Since the robustness of DNNs to adversarial examples is a critical feature, \emph{defenses} that target to increase robustness against adversarial example are urgently considered and can be classified into three main categories: input transformations~\cite{dziugaite2016study,lu2017no,meng2017magnet}, adversarial training~\cite{szegedy2013intriguing} and gradient masking~\cite{papernot2017practical,zheng2016improving}. In addition to defense, \emph{detection} of adversarial examples before they are fed into the networks is another approach to resist attacks, such as MagNet~\cite{meng2017magnet} and steganalysis based detection~\cite{liu2019detection}.

The popularity of 3D sensors such as LiDAR and RGBD cameras draws many research concerns with 3D vision. An increasing number of accessible data motivates data-driven deep learning methods practical to be used in many areas, including autopilot~\cite{qi2018frustum,zhou2018voxelnet}, robotics~\cite{jaremo2018density,deng2018ppfnet} and graphics~\cite{yan2016perspective,kato2018neural,wang2018pixel2mesh}. In particular, point cloud is one of the most natural data structures to represent the 3D geometry. After the difficult problem of irregular data format was addressed by DeepSets ~\cite{zaheer2017deep}, PointNet ~\cite{charles2017pointnet} and its variants ~\cite{qi2017pointnet++,wang2018dynamic}, point cloud data can be directly processed by DNNs, and has become a promising data structure for 3D computer vision tasks. Hua \etal ~\cite{hua2018pointwise} propose a pointwise convolution operator that can output features at each point, which can offer competitive accuracy while being easy to implement. Yang \etal~\cite{yang2018realistic} construct losses based on mesh shape and texture to generate adversarial examples, which aim to project the optimized ``adversarial meshes'' to 2D with a photorealistic renderer, and still able to mislead different DNNs. Xiang \etal~\cite{xiang2019generating} attack point clouds built upon C\&W loss and point cloud-specific perturbation metric with high success rate. Zheng \etal~\cite{zheng2018learning} propose a malicious point-dropping method to generate adversarial point clouds based on learning a saliency map for a whole point cloud, which assigns each point a score reflecting its contribution to the model-recognition loss. Liu \etal~\cite{liu2019extending} propose several iterative gradient based attack methods and input restoration based defenses. Yang \etal~\cite{yang2019adversarial} propose point-detach strategy utilizing the critical point property to attack neural network based classification system, and introduce several perturbation defenses.

Adversarial examples do well in 3D point cloud classification, and probably cause inconvenient issues even security problems. Consequently, research on defense of 3D point cloud adversarial example is in urgent need. Based on the above reasoning, in this paper, we propose a defense method against adversarial point cloud by training a Denoiser and UPsampler Network (DUP-Net) to mitigate adversarial effects. As far as we know, this is the first work that demonstrates the effectiveness of point-dropping and point-adding operations at inference time on mitigating adversarial effects on the 3D dataset, \eg, ModelNet40. We summarize the key contributions of our work as follows:
\begin{itemize}
\item We present two new defense modules to mitigate adversarial point clouds, which have better performance compared with baseline methods.
\item The nondifferentiability property of denoise layer, statistical outlier removal, is utilized to defend the adversarial white-box attacks.
\item The upsampler network can be trained on a small dataset and defends well against adversarial attacks generated from other point cloud datasets.
\end{itemize}

We conduct comprehensive experiments to test the effectiveness of our defense method against point shifting/dropping/adding attacks from ~\cite{xiang2019generating,yu2018pu}. The results in Section~\ref{sect:02} demonstrate that the proposed DUP-Net can significantly mitigate adversarial effects.

\section{Related Work}
\subsection{Point Clouds and PointNet}
A point cloud is a set of points which are sampled from object surfaces. Consider a 3D point cloud with $n$ points, denoted by $\textbf{X}=\{\textbf{x}_i|i=1,2,...,n\}$, where each point $\textbf{x}_i$ is a vector of its $\emph{xyz}$ coordinates.
PointNet ~\cite{qi2017pointnet} and its variants~\cite{qi2017pointnet++} proposed by Qi \etal exploit a single symmetric function, max pooling, to reduce the unordered and dimensionality-flexible input data to a fixed-length global feature vector and enable end-to-end neural network learning. They demonstrate the robustness of PointNet and introduce the concept of critical points and upper bounds. The points sets laying between critical points and upper bounds yield the same global features, and thus PointNet is robust to missing points and random perturbation.

\subsection{Existing methods for adversarial attacks}

\begin{figure*}
\begin{center}
{\centering\includegraphics[width=0.85\linewidth]{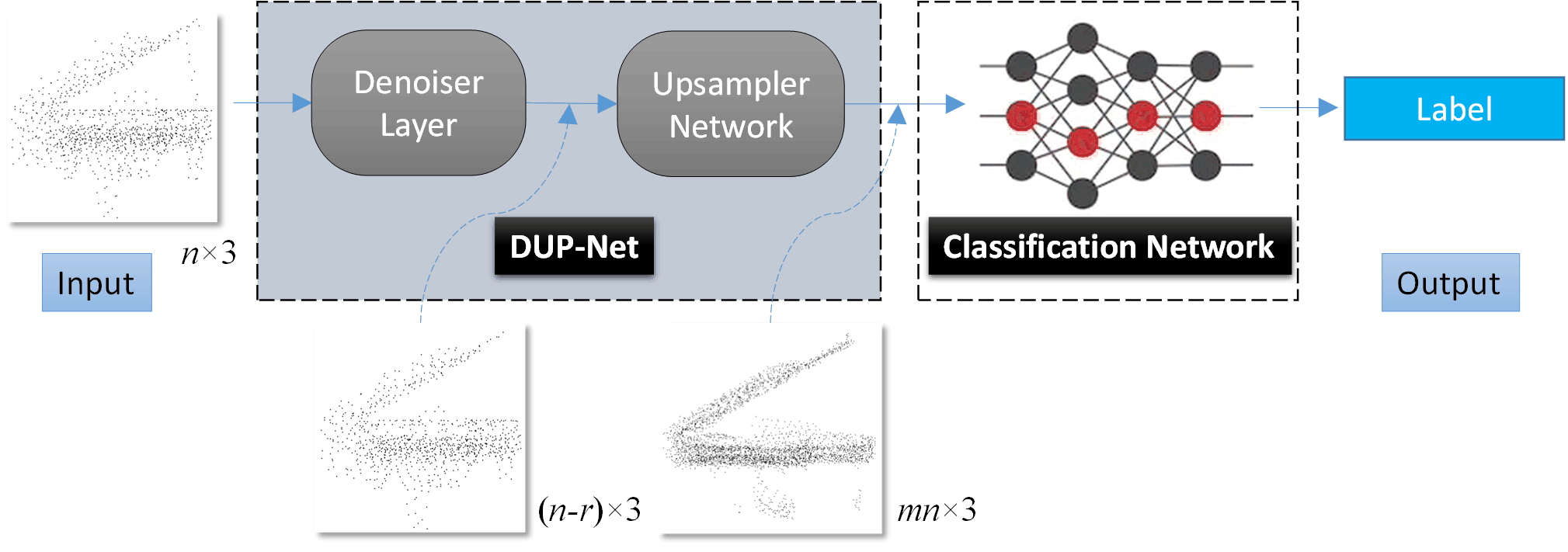}
}\\
\caption{The pipeline of our DUP-Net defense method. The input point cloud is first denoised by a statistical outlier removal layer and then upsampled by a pretrained upsampling neural network. The preprocessed point cloud is then fed into the classification neural network.}\label{fig:01}
\end{center}
\end{figure*}

\textbf{Point Shifting.}
Xiang \etal~\cite{xiang2019generating} propose C\&W framework~\cite{carlini2017towards} based unnoticeable adversarial point clouds by point perturbation. C\&W is an optimization-based attack that combines a differentiable surrogate for the classification accuracy of the model. It generates adversarial examples by solving the following optimization problem:

\begin{equation}
\label{eqn:01}
\begin{aligned}
\min_{\delta}&D(\textbf{X}, \textbf{X}+\delta)+c\cdot f(\textbf{X}+\delta)\\
&s.t. \quad \textbf{X}+\delta\in[0,1]^n
\end{aligned}
\end{equation}
This attack seeks for the solution of both acquiring the smallest perturbation $D$ and impelling the network to classify the adversarial example incorrectly. For an untargeted attack, $f(\textbf{X})$ is the loss function to measure the distance between the input and the adversarial object, as defined by:

\begin{equation}
\label{eqn:02}
f(\textbf{X})=\max (Z(\textbf{X})_{true}-\max_{i\neq true}\{Z(\textbf{X})_i\}, -\kappa)
\end{equation}
where $\kappa$ denotes a margin for regulating model transferability and perturbation degree, and $Z(\textbf{X})$ is the logit vector.

Xiang \etal shift existing points negligibly and adopt different perturbation metrics $D(\textbf{X}, \textbf{X}')$ ($l_p$ norm, Hausdorff distance and Chamfer measurement), where $\textbf{X}'$ stands for adversarial point cloud.
Liu \etal~\cite{liu2019extending} extend fast/iterative gradient method by constraining the perturbation magnitude onto the surface of an epsilon ball in different dimensions.

\textbf{Point Adding.}
Xiang \etal~\cite{xiang2019generating} also propose points adding based attacks with C\&W and Hausdorff/Chamfer measurements. Because directly adding points to the unconstrained 3D space is infeasible due to the large search space, they propose an initialize-and-shift method to find appropriate position for each added point. Besides, for adversarial clusters, they  minimize the radius of the generated cluster to make it attached to the surface of original object.
Yang \etal~\cite{yang2019adversarial} develop a variant of one-pixel attack~\cite{su2019one} using pointwise gradient method to only update the attached points without changing the original point cloud.

\textbf{Point Dropping.}
For any point cloud $\textbf{X}$, Qi \etal~\cite{qi2017pointnet} prove that there exists a subset $\textbf{C}\subseteq\textbf{X}$, namely critical subset, which determines all the max pooled features $\max\limits_{\textbf{x}_i\in\textbf{X}}\{\textbf{h}(\textbf{x}_i)\}$, and thus the output of PointNet, which is only applicable to network structures similar to $\gamma\circ\max\limits_{\textbf{x}_i\in\textbf{X}}\{\textbf{h}(\textbf{x}_i)\}$. Visually, $\textbf{C}$ usually distributes evenly along the skeleton of $\textbf{X}$. In this sense, dropping points in $\textbf{C}$ can also generate adversarial point clouds. Zheng \etal~\cite{zheng2018learning} propose point dropping based attack by first constructing the saliency map:
\begin{equation}
\label{eqn:0z}
s_i=-r_i^{\alpha}r_i\frac{\partial{\mathcal{L}}}{\partial{r_i}},
\end{equation}
where $r_i^{\alpha}\frac{\partial{\mathcal{L}}}{\partial{r_i}}=(\textbf{x}_i-\textbf{x}_c)\cdot\textbf{g}_i$ (inner product), $\textbf{x}_c$ is the cloud center and $\textbf{g}_i$ is the gradient under orthogonal coordinates $\textbf{g}_i=\nabla_{\textbf{x}_i}\mathcal{L}(\textbf{X},y;\bm{\theta})$, and $\alpha$ is a rescaling hyperparameter. Points with $n/T$ lowest $s_i$ are dropped, and the operation are iterative executed for $T$ loops. Yang \etal~\cite{yang2019adversarial} develop a point-detach strategy similar to \cite{zheng2018learning}, which utilizes the critical point property to iteratively detach the most important point to confuse the attacked network.

\subsection{Existing Methods for defenses}

\textbf{Adversarial Training.}
Adversarial training~\cite{goodfellow2014explaining,kurakin2016adversarial,tramer2017ensemble} is one of the most investigated defenses against adversarial attacks, which augment the training set with adversarial examples to increase the model's robustness against a specific attack. Adversarial training improves the classification accuracy of the target model on adversarial examples. 

\textbf{Simple Random Sampling.}
In statistics, a simple random sampling, or shortly SRS, is a subset of individuals chosen from a larger set, where each sample is chosen randomly with the same probability. For $\textbf{X}$ containing $n$ points, we randomly sample $r$ ($r<n$) points from $\textbf{X}$ by
\begin{equation}
\label{eqn:0x}
\mathcal{P}_i(\textbf{X})=\{\mathds{1}_{\textbf{x}}|\textbf{x}\in\textbf{X}, \mathds{1}_{\textbf{x}}\sim Ber(0.5)\},
\end{equation}
where $\textbf{x}$ is sampled from $Bernoulli(0.5)$ distribution to indicate the existence of point $\textbf{x}$ in the post-sampled set.

As described in \cite{yang2019adversarial}, Gaussian noising and quantification are another two basic defenses for point clouds, which are similar to defenses for image adversarial examples.

\section{Defenses against Adversarial Point Cloud}

The goal of defense on 3D point clouds is to build a network that is robust to adversarial examples, \ie, it can classify adversarial point clouds correctly with little performance loss on clean point clouds. Formally, given a classification model $f$ and an input $\tilde{\textbf{X}}$, which may either be an original input $\textbf{X}$, or an adversarial input $\textbf{X}'$, the goal of a defense method is to either augment data to train a robust $f'$ such that $f'(\tilde{\textbf{X}})=f(\textbf{X})$, or transform $\tilde{\textbf{X}}$ by a transformation $\mathcal{T}$ such that $f(\mathcal{T}(\tilde{\textbf{X}}))=f(\textbf{X})$.

Towards this goal, we propose a method formed by a denoiser and an upsampler, as shown in Figure~\ref{fig:01}, which adds an outlier-removal layer and an upsampling network to the front of the classification networks, to realize network robustness. These layers are designed in the context of point cloud classification on ModelNet40~\cite{wu20153d} dataset and are used in conjunction with a trained classifier $f$. 
The defense function is denoted as $D: \textbf{X}'\rightarrow \hat{\textbf{X}}$, where $\hat{\textbf{X}}$ denotes the denoised and upsampled point cloud. Inspired from upsampling network for generating a denser and uniform set of points~\cite{yu2018pu}, we define the loss function as
\begin{equation}
\label{eqn:0x}
\mathcal{L}(\textbf{X},\hat{\textbf{X}})=\mathcal{L}_{rec}(\textbf{X},\hat{\textbf{X}})+\beta\mathcal{L}_{rep}(\textbf{X},\hat{\textbf{X}})+\gamma\|\bm{\theta}\|_2^2,
\end{equation}
where $\mathcal{L}_{rec}$ is the reconstruction loss and $\mathcal{L}_{rep}$ the repulsion loss. $\bm{\theta}$ indicates the network parameters, $\beta$ balances the reconstruction loss and repulsion loss, and $\gamma$ denotes the multiplier of weight decay.

\subsection{Statistical Outlier Removal (SOR)}
\label{sect:01}

Since there exists outliers in raw point cloud data produced by 3D scanners, Rusu \etal~\cite{rusu2008towards} propose statistical outlier removal method (SOR for short) which corrects these irregularities by computing the mean $\mu$ and standard deviation $\sigma$ of nearest neighbor distances, and trim the points which fall outside the $\mu\pm\alpha\cdot \sigma$, where $\alpha$ depends on the size of the analyzed neighborhood.

We use $k$-nearest neighbors ($k$NN) for outlier removal. Specifically, the $k$NN point set of each point $\textbf{x}_i$ of point cloud $\textbf{X}$ is defined as $knn(\textbf{x}_i, k)$. Then the average distance $d_i$ that each point $\textbf{x}_i$ has to its $k$NN is denoted by
\begin{equation}
\label{eqn:06}
d_i=\frac{1}{k}\sum_{\textbf{x}_j\in knn(\textbf{x}_i, k)}\|\textbf{x}_i-\textbf{x}_j\|_2, \quad i=1,\ldots,n.
\end{equation}

The mean and standard deviation of all these distances are computed to determine a distance threshold:
\begin{equation}
\label{eqn:07}
\bar{d}=\frac{1}{n}\sum_i^n d_i, \quad i=1,\ldots,n,
\end{equation}

\begin{equation}
\label{eqn:08}
\sigma=\sqrt{\frac{1}{n}\sum_i^n(d_i-\bar{d}_i)^2}.
\end{equation}

We trim the points which fall outside the $\mu\pm\alpha\cdot \sigma$, and the manicured point set $\textbf{X}'$ is acquired by
\begin{equation}
\label{eqn:09}
\textbf{X}'=\{\textbf{x}_i|d_i<\bar{d}+\alpha\cdot \sigma\}.
\end{equation}

We explore the relationships between outliers and adversarial points to explain why SOR is effective for defending C\&W based attacks. Despite that the attackers successfully fool the classification network, there is always a certain percentage of points that inevitably become abnormal points and are then captured and dropped by SOR.
C\&W attack makes some points off the manifold of point cloud surface, and such outliers mostly mislead the classification performance. Therefore, the more outliers removed by preprocessing layer, the better the defense ability against adversarial examples. Here, we denote the percentage of adversarial points in the removed point set by
\begin{equation}
\label{eqn:10}
p=\frac{|\textbf{X}_{adv}\cap(\textbf{X}-\textbf{X}')|}{|\textbf{X}|-|\textbf{X}'|},
\end{equation}
where $\textbf{X}_{adv}$ is the set of adversarial points which is defined differently \wrt diverse adversarial distortion constraints. For a $l_2$ loss, $\textbf{X}_{adv}$ is defined by
\begin{equation}
\label{eqn:11}
\textbf{X}_{adv}=\{\textbf{x}'_i|\|\textbf{x}_i-\textbf{x}'_i\|_2>T(\textbf{X},\textbf{X}',\epsilon)\},
\end{equation}
where $\textbf{x}'_i\in \textbf{X}'$ and $T(\textbf{X},\textbf{X}',\epsilon)$ is the threshold of $l_2$ norm of each paired points controlled by $\epsilon$ the ratio of points that are considered as adversarial points. For Hausdorff or Chamfer based loss, $\textbf{X}_{adv}$ is defined by
\begin{equation}
\label{eqn:12}
\textbf{X}_{adv}=\{\textbf{x}'_i|\min_{\textbf{m}\in \textbf{X}}\|\textbf{m}-\textbf{x}'_i\|_2>T(\textbf{X},\textbf{X}',\epsilon)\},
\end{equation}
where $\textbf{x}'_i\in \textbf{X}'$ and $T(\textbf{X},\textbf{X}',\epsilon)$ is the threshold of Hausdorff/Chamfer distance between each point from $\textbf{X}'$ and point set $\textbf{X}$ controlled by $\epsilon$.

\begin{figure}
\begin{center}
\includegraphics[width=1.08\linewidth]{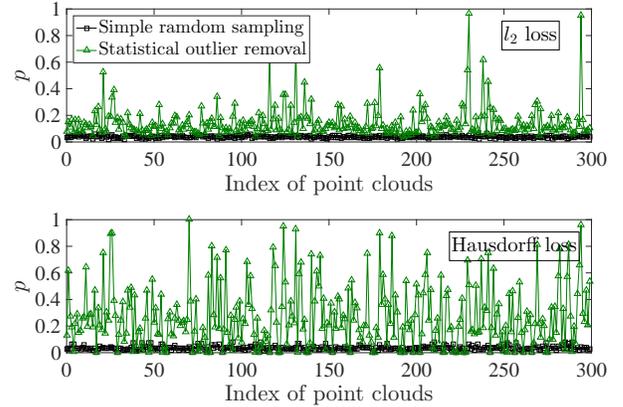}
\\
\caption{Comparison of $p_{\textbf{SOR}}$ and $p_{\textbf{SRS}}$ under $l_2$ and Hausdorff loss based targeted adversarial examples on PointNet network, respectively. The ratio $\epsilon$ is set with $0.04$. }\label{fig:02}
\end{center}
\end{figure}

\begin{figure}
\begin{center}
{\centering\includegraphics[width=1.00\linewidth]{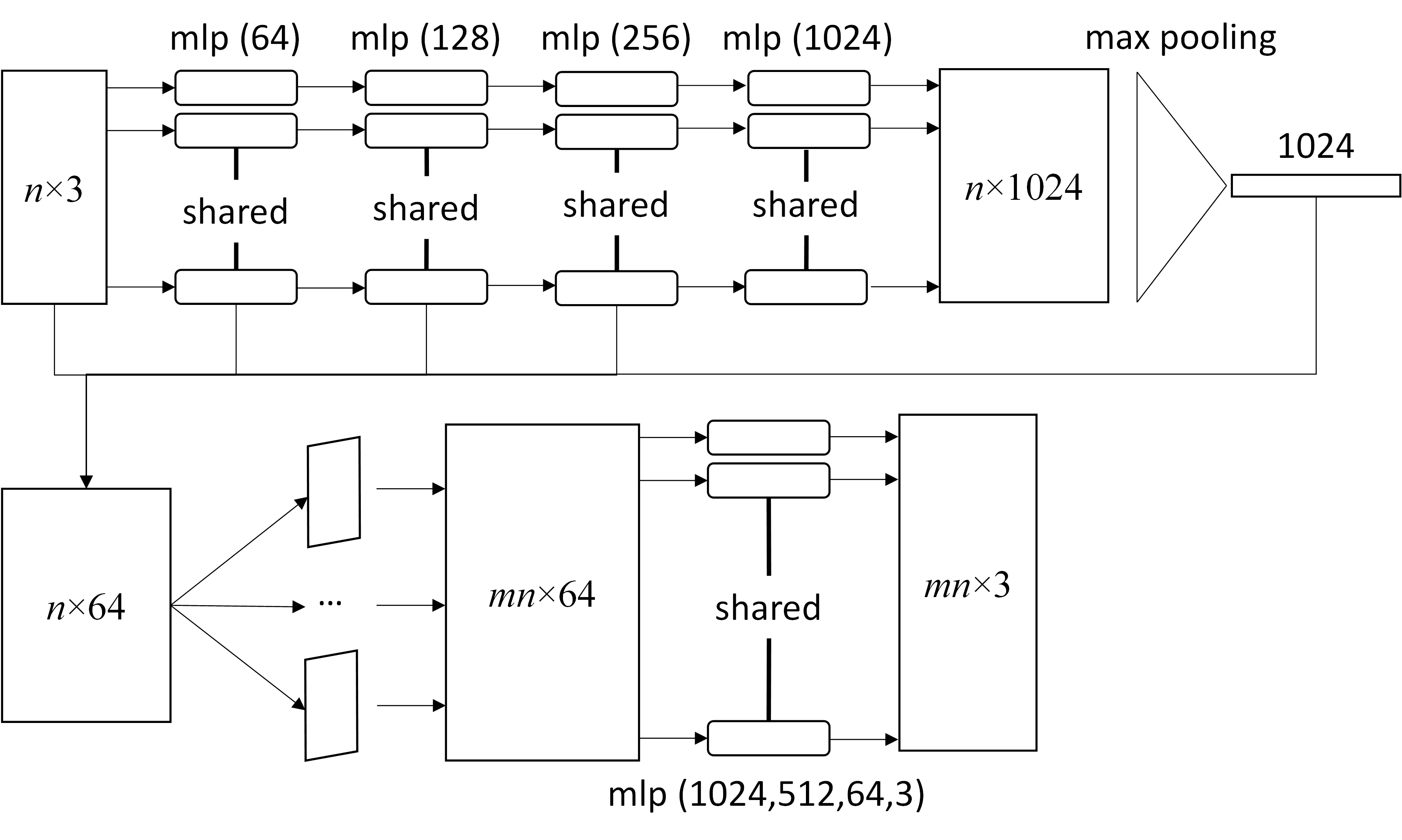}
}\\
\caption{The network architecture of point cloud upsampling network (PU-Net).}\label{fig:03}
\end{center}
\end{figure}

By Equation~(\ref{eqn:10}), we acquire the percentage of adversarial points $p$ of SOR and SRS methods, and denote them by $p_{\textbf{SOR}}$ and $p_{\textbf{SRS}}$ respectively. It is expected that $p_{\textbf{SOR}}>p_{\textbf{SRS}}$ since SOR scheme recognizes outliers as adversarial points in a statistical pattern rather than random guess as SRS does. We choose 300 point clouds as test examples to verify the above inference, as shown in Figure~\ref{fig:02}. Most of $p_{\textbf{SOR}}$ of point clouds are larger than $p_{\textbf{SRS}}$, implying that SOR drops more adversarial points than SRS. 
Thus SOR has a better ability of defense against adversarial point clouds than SRS.

\subsection{Upsampling Network}
Our goal is to defend a classification model $f$ against the perturbed point clouds generated by an adversary. Our approach is motivated by the manifold assumption \cite{zhu2009introduction}, which postulates that natural images or point clouds lie on low-dimensional manifolds. The perturbed point clouds are known to lie off the low-dimensional manifold of natural point clouds, which is approximated by deep networks. 
We propose to use point cloud upsampling to remap off-the-manifold adversarial samples on to the natural manifold to reconstruct surface. In this manner, robustness against adversarial perturbations is achieved by enhancing the visual quality of point clouds. This approach offers significant advantages over other defense mechanisms that truncate critical information to achieve robustness.

Since these perturbations are generally missing critical points from skeletons from point clouds, we use a point cloud upsampling network that output a denser point cloud that follows the underlying surface of the target object while being uniform in distribution. The network considered in this work is the Point Cloud Upsampling Network (PU-Net) \cite{yu2018pu}, which learns geometry semantics of point-based patches from 3D models, and the architecture is illustrated in Figure~\ref{fig:03}. To address the uncertain presence of the varying part, the minimizer of Chamfer distance (CD) distributes some points outside the main body at the correct locations; while the minimizer of Earth Mover's distance (EMD) is considerably distorted~\cite{fan2017point}. We also try EMD \cite{fan2017point} loss to observe the defense performance:
\begin{equation}
\label{eqn:1t}
\mathcal{L}_{rec}=D(\textbf{X}, \hat{\textbf{X}})=\frac{1}{\|\hat{\textbf{X}}\|_0}\sum_{\textbf{x}'\in \hat{\textbf{X}}}\min_{\textbf{x}\in \textbf{X}}\|\textbf{x}-\textbf{x}'\|_2^2.
\end{equation}
The total loss function combines the reconstruction loss $\mathcal{L}_{rec}$ and repulsion loss $\mathcal{L}_{rep}$ proposed in \cite{fan2017point}. In contrast, we use a simple upsampling method~\cite{alexa2003computing}, which interpolates points at vertices of a Voronoi diagram for comparative trial.

In summary, we have given a formal definition of the proposed defense in Algorithm~\ref{alg:01}.

\begin{algorithm}[t]
\caption{Denoise and upsample points as defense}
\label{alg:01}
\hspace*{0.02in} {\bf Input:}
Point cloud $\textbf{X}$, nearest neighbor number $k$, outlier truncation parameter $\alpha$ and network parameter $\bm{\theta}$\\
\hspace*{0.02in} {\bf Output:}
Prediction label $l$
\begin{algorithmic}[1]
\State Initialize $\textbf{X}'=\varnothing$
\State Compute the average distance $d_i$ that each point $\textbf{x}_i$ has to its nearest $k$ neighbors by Equation~(\ref{eqn:06})
\State Compute the mean $\bar{d}$ and standard deviation $\sigma$ of all these distances by Equation~(\ref{eqn:07}) and (\ref{eqn:08})
\For{$i\leftarrow 0$ to $n$}
　　\If{the average distance $d_i<\bar{d}+\alpha\cdot \sigma$}
　　　　\State $\textbf{X}'=\textbf{X}'\cup\textbf{x}_i$ ($\textbf{x}_i\in\textbf{X}$)
　　\EndIf
\EndFor
\State The upsampled point clouds $\hat{\textbf{X}}$ is generated by feeding $\textbf{X}'$ into upsampling network
\State $\hat{\textbf{X}}$ is fed into classification network $f(\hat{\textbf{X}})$, and it outputs the prediction label $l$
\State \Return Prediction label $l$
\end{algorithmic}
\end{algorithm}

\begin{figure*}
\begin{center}
\centering
\subfloat[]{
	\label{fig:subfig_a}
	\includegraphics[width=0.32\linewidth]{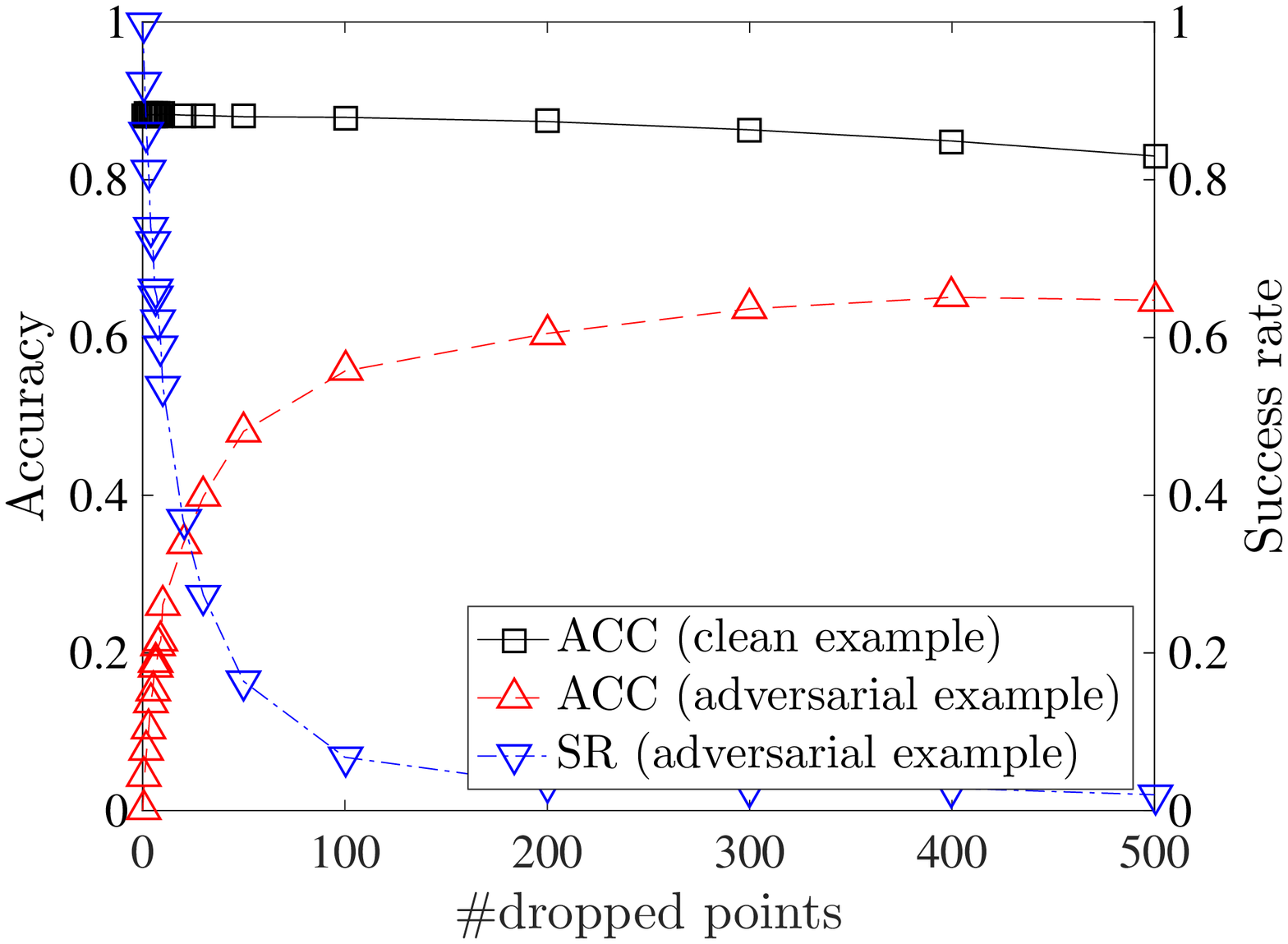}
}
\centering
\subfloat[]{
	\label{fig:subfig_b}
	\includegraphics[width=0.32\linewidth]{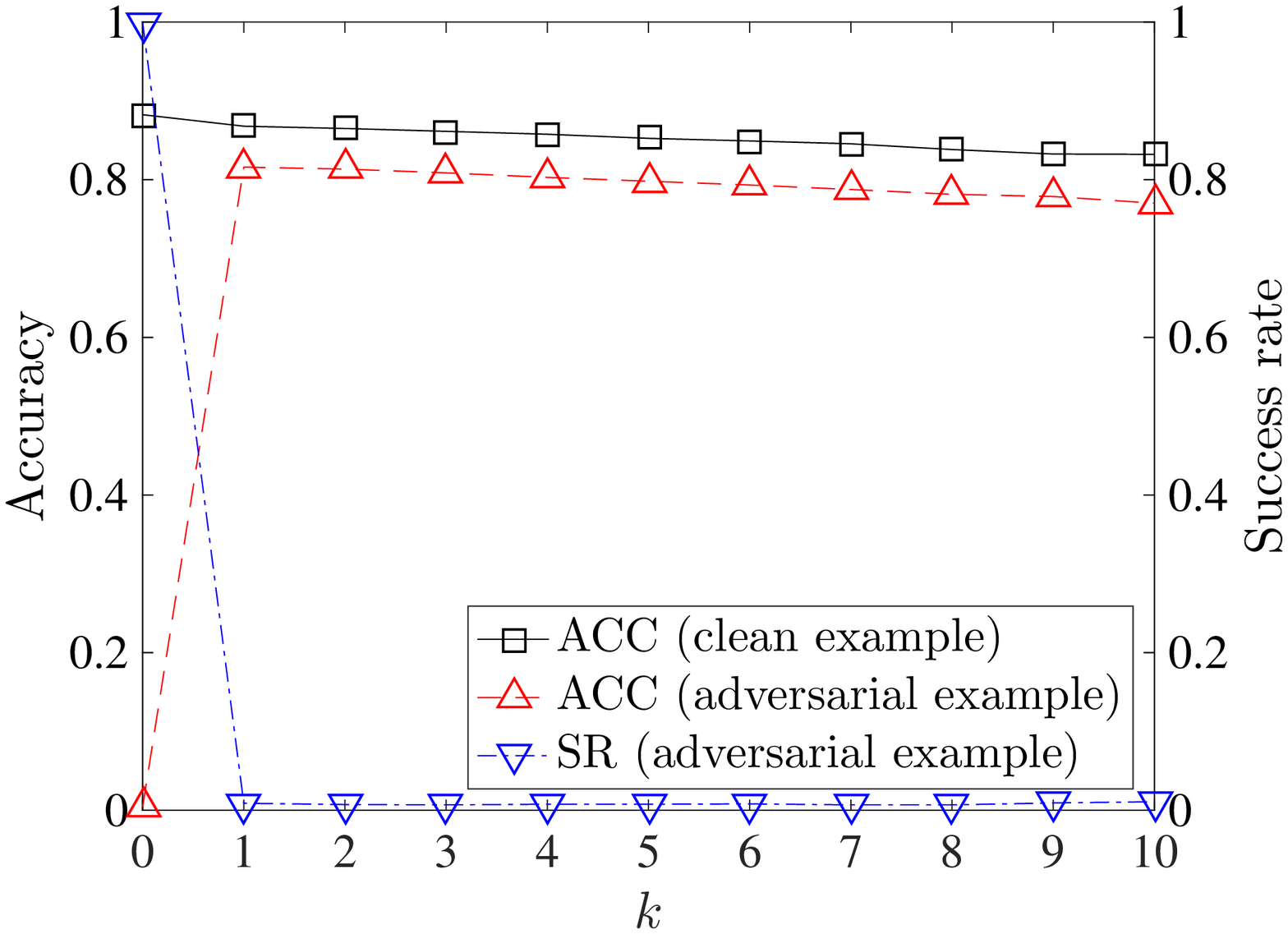}
}
\centering
\subfloat[]{
	\label{fig:subfig_c}
	\includegraphics[width=0.32\linewidth]{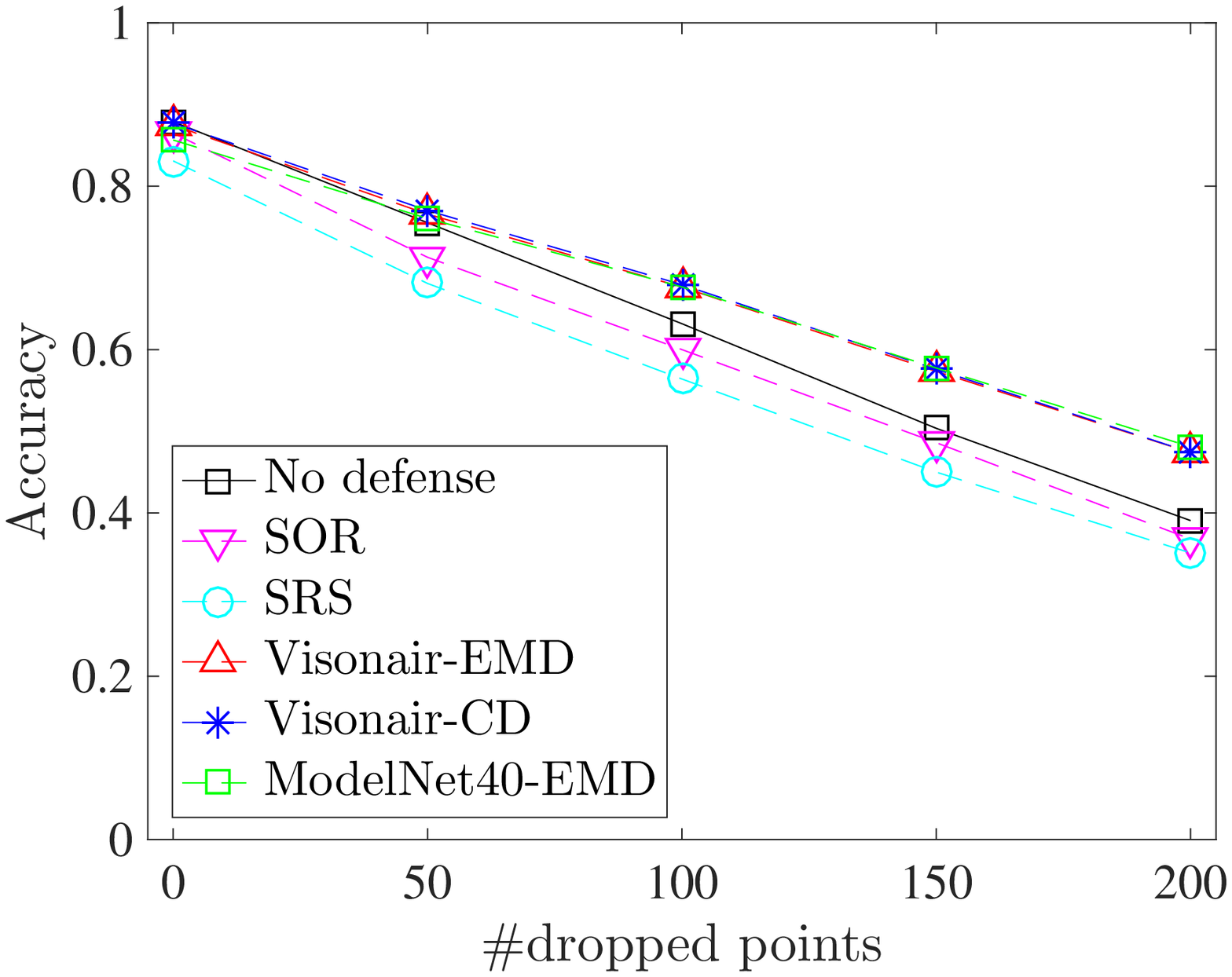}
}\\
\caption{(a) SRS defense performance of clean and targeted adversarial point clouds on PointNet using C\&W and shifting loss; (b) SOR defense performance of clean and targeted adversarial point clouds on PointNet using C\&W and shifting loss under $\alpha=1.1$; (c) Defense performance of adversarial point clouds on PointNet with or without defense. }
\label{fig:04}
\end{center}
\end{figure*}

\section{Experiments}
\label{sect:02}
\subsection{Experimental Setup}

\begin{table*}
\begin{center}
\begin{tabular}{c||c|c|c|c|c}
\hline
\textbf{Models} & \textbf{Target}~\cite{charles2017pointnet} & \tabincell{c}{\textbf{Defense}\\ \textbf{(SRS)}} & \tabincell{c}{\textbf{Defense (SOR)}\\(ours)} & \tabincell{c}{\textbf{Defense (PU-Net)}\\(ours)} & \tabincell{c}{\textbf{Defense (DUP-Net)}\\(ours)}\\
\hline
Clean point cloud                                        &  \textbf{88.3\%} & 83.0\% & 86.5\% & 87.5\% & 86.3\%\\
Adv (C\&W $+$ $l_2$ loss)~\cite{xiang2019generating}     &   0.7\% & 64.7\% & 81.4\% & 23.9\% & \textbf{84.5}\%\\
Adv (C\&W $+$ Hausdorff loss)~\cite{xiang2019generating} &  12.7\% & 58.8\% & 59.8\% & 17.6\% & \textbf{62.7\%}\\
Adv (C\&W $+$ Chamfer loss)~\cite{xiang2019generating}   &  11.8\% & 59.5\% & 59.1\% & 18.0\% & \textbf{61.4\%}\\
Adv (C\&W $+$ 3 clusters)~\cite{xiang2019generating}   &  0.7\% & \textbf{92.0\%} & - & - & 87.6\%\\
Adv (C\&W $+$ 3 objects)~\cite{xiang2019generating}   &  2.7\% & \textbf{92.4\%} & - & - & 68.4\%\\
Adv (dropping 50 points)~\cite{zheng2018learning}        &  75.5\% & 68.1\% & 71.3\% & \textbf{76.1\%} & 73.9\%\\
Adv (dropping 100 points)~\cite{zheng2018learning}       &  63.2\% & 56.4\% & 60.0\% & \textbf{67.7\%} & 64.3\%\\
Adv (dropping 150 points)~\cite{zheng2018learning}       &  50.4\% & 45.0\% & 48.6\% & \textbf{57.7\%} & 54.4\%\\
Adv (dropping 200 points)~\cite{zheng2018learning}       &  39.1\% & 35.1\% & 36.8\% & \textbf{48.1\%} & 43.7\%\\
\hline
\end{tabular}
\end{center}
\caption{Classification accuracy under the white-box attack on PointNet. For the SRS defense model, number of random dropped points is 500 and for SOR defense model, $k=2$ and $\alpha=1.1$ are set as hyperparameters. }
\label{tab:01}
\end{table*}

\textbf{Dataset.}
We utilize ModelNet40~\cite{wu20153d} datasets to test our proposed DUP-Net, which contains 12,311 CAD models from 40 object categories, where 9,843 objects are used for training and the other 2,468 for testing. As done by Qi \etal, before generating adversarial point clouds using~\cite{yu2018pu,xiang2019generating} methods, first uniformly sample 1,024 points from the surface of each object and rescale them into a unit cube.
We also use Visionair dataset collected by \cite{yu2018pu} for training DUP-Net, which contains 60 different models from the Visionair repository~\cite{Visionair}, ranging from smooth non-rigid and steep rigid objects.

\textbf{Networks and Implementation Details.}
We use PointNet~\cite{qi2017pointnet} and PointNet++~\cite{qi2017pointnet++} as targeted classification networks and train the models using default settings. 
To train the proposed DUP-Net, for ModelNet40, the upsampled point number is $2048$ and the upsampling rate is 2; for Visionair dataset, the number is 4096 and upsampling rate is 4. Each input sample has $n=1024$ points for both training procedures. The balancing weights $\beta$ and $\gamma$ are set as 0.01 and $10^{-5}$, respectively. The implementation is based on TensorFlow. For the optimization, we train the network for 120 epochs using the Adam~\cite{kingma2014adam} algorithm with a minibatch size of 28 and a learning rate of 0.001. 

\textbf{Attacks Evaluations.}
The attackers first generate adversarial examples using the untargeted/targeted models and then evaluate the classification accuracy of these generated adversarial examples on the target and defense models. Low accuracy of the untargeted/targeted model indicates that the attack is successful, and high accuracy of the defense model indicates that the defense is effective.

\subsection{Ablation Study}
Our proposed defense first deploys the SOR layer, which aims to minimize the effect of outlier points perturbations, followed by an upsampling network to selectively introduce missing points into a point cloud and recover off-the-manifold attacked point clouds.

\textbf{SOR as Defense.}
We compare the detection accuracy and attack success rate of targeted attacks of proposed SOR defense with baseline SRS for C\&W and shifting~\cite{xiang2019generating} based attacks and dropping~\cite{zheng2018learning} based attacks on PointNet. 
Gaussian noising and quantification are not considered as defense because they will degrade accuracy of clean samples.

\begin{table}
\begin{center}
\begin{tabular}{c||c|c|c}
\hline
\textbf{Models} & CW $l_2$ & CW Hausd & Drop 200 \\
\hline
\textbf{Target}~\cite{qi2017pointnet++}  & 0\%  & 28.1\% & 56.4\% \\
\textbf{Defense (SRS)} & 66.7\%  & 51.7\% & 47.3\% \\
\textbf{Defense (DUP-Net)} & \textbf{75.7\%}  & \textbf{54.1\%} & \textbf{61.9\%} \\
\hline
\end{tabular}
\end{center}
\caption{\small{Comparison of classification accuracy using SRS and proposed DUP-Net under PointNet++ network. }}
\label{tab:03}
\end{table}

\begin{table*}
\begin{center}
\begin{tabular}{c||c|c|c|c|c|c}
\hline
Networks & Task & \tabincell{c}{Shifting\\($l_2$)~\cite{xiang2019generating}} & \tabincell{c}{Adding\\(Hausdorff)~\cite{xiang2019generating}} & \tabincell{c}{Adding\\(Chamfer)~\cite{xiang2019generating}} & \tabincell{c}{Dropping 100\\points~\cite{zheng2018learning}} & \tabincell{c}{Dropping 200\\points~\cite{zheng2018learning}}\\
\hline
 \textbf{PointNet}~\cite{qi2017pointnet}  & Target  & 0.7\%  & 12.7\% & 11.8\%     & 63.2\% & 39.1\%  \\
\hline
\multirow{3}{*}{\textbf{PointNet++}~\cite{qi2017pointnet++}} & Target  & \textbf{89.5\%} & 52.9\%  & 51.0\% & \textbf{82.4\%} & \textbf{75.6\%}\\
&\tabincell{c}{Defense (SRS)}  & 82.9\% & \textbf{59.6\%}  & \textbf{58.3\%} & 70.3\% & 54.5\%\\
&Defense (DUP-Net)  & 84.3\% & 48.3\%  & 48.5\% & 75.2\% & 67.3\%\\
\hline
\multirow{3}{*}{\textbf{DGCNN}~\cite{wang2018dynamic}} & Target & \textbf{91.2\%}  & \textbf{51.4\%}  & \textbf{50.8\%} & \textbf{75.5\%} & \textbf{67.9\%}\\
& Defense (SRS)  & 68.2\% & 38.2\%  & 37.1\% & 72.2\% & 54.9\%\\
& Defense (DUP-Net)  & 40.7\% & 25.5\%  & 26.7\% & 32.0\% & 26.7\%\\
\hline
\end{tabular}
\end{center}
\caption{Black-box attacks and defenses: accuracy of targeted C\&W and shifting and adding based adversarial point clouds and saliency map and points dropping based adversarial point clouds generated from PointNet on other classification networks~\cite{qi2017pointnet++,wang2018dynamic,hua2018pointwise} with or without defense. }
\label{tab:02}
\end{table*}

As shown in Figure~\ref{fig:04}a, for SRS baseline, as dropped points increases, the attack success rate drops dramatically, and the accuracy gradually increases with its maximum 65.1\%; for clean examples, the accuracy is monotonically decreasing.
The tendency of three curves can be explained: the attacks search the entire point cloud space for adversarial perturbations without considering the location of the point cloud content, which is contrary to the classification models that pay attention to object shapes~\cite{yosinski2015understanding}. Therefore, dropping a few points with SRS erases the artifact bothered by adversarial perturbation, which promotes defense of adversarial point clouds. The structure of point cloud is still preserved with a few points dropped; when more random sampled points are dropped, the shape of the point cloud deteriorates and thus the classification accuracy degrades.

As shown in Figure~\ref{fig:04}b, the SOR operation comprises two influential factors, $k$ the number of neighbor points and $\alpha$ the percentage of points that are regarded as outliers. When $k=0$, the $k$NN point set only contains the point itself, thus the statistical removal defense is noneffective; when $k\geq 1$, the defense works. When $k=2$ and $\alpha=1.1$, the accuracy of clean point clouds and adversarial examples are 86.5\% and 81.4\% respectively. Compared to SRS defense with its best accuracy of adversarial examples 65.1\%, SOR has a substantial increase of 16.3\% on performance. Similar results can be obtained on defenses of untargeted attacks and Hausdorff loss based attacks.

For the vanilla attack scenario, the attackers are not aware of the points-removal layer, and directly use the original networks as the target model to generate adversarial examples. For reading convenience, we coin a new acronym  ``adv'' standing for ``adversarial point clouds'' in tables. As shown in Table~\ref{tab:01}, the points-removal layer can mitigate the adversarial effects on C\&W methods significantly.
As for $l_2$ metric, SOR has the best performance with 81.4\% accuracy. For Hausdorff and Chamfer loss metrics, SOR and SRS have similar results with lower accuracy.

To validate whether differentiable point removal layer is able to simulate non-differentiable layer, We train the modified PointNet with dropping probability $p=0.5$ before the max pooling layer, and conduct white-box attacks. The classification accuracy of C\&W $l_2$, C\&W Hausdorff and drop 200 attacks are 0\%, 0\% and 54.5\%, respectively,  which implies that the differentiable random point removal cannot simulate non-differentiate layer well.

\textbf{PU-Net as Defense.}
Here we investigate the individual impact of PU-Net module towards defending adversarial attacks. Since C\&W based attack can be defended by the proposed SOR layer, in this section we only consider saliency point dropping based attack proposed by Zheng~\etal~\cite{zheng2018learning}. We perform extensive experiments on three upsampler networks: pretrained PU-Net model from~\cite{yu2018pu} (Visonair-EMD), PU-Net with Chamfer distance loss trained by Visonair (Visonair-CD) and PU-Net with Earth Mover's distance trained by ModelNet40 (ModelNet40-EMD), as shown in Figure~\ref{fig:04}c. The results demonstrate that, upsampler network helps adversarial examples filling missing points that are critical for classification, especially, when dropped number is 200, the defense has nearly 9\% increase. Besides, the upsampler network can be trained on a small dataset and defends well against adversarial attacks generated from other point cloud datasets with fine generalization ability. We find that the performance of the model trained by CD loss is similar with that trained by EMD loss \wrt adversarial example defense, which means that the selection of loss function does not affect classification accuracy. SOR and SRS both deteriorate the defense performance, because the dropping attacks visually damage the local shapes of point clouds, and the SOR/SRS operation further amplifies the distortion.

We also compare PU-Net with a simple upsampling method~\cite{alexa2003computing}, which interpolates points at vertices of a Voronoi diagram. Experiments show that PU-Net performs much better than~\cite{alexa2003computing} by 8\% when attacked by ``Drop 200".

\subsection{Evaluation Results of DUP-Net}
The last column in Table~\ref{tab:01} shows the overall defense accuracy of proposed DUP-Net against different attacks (C\&W based points-shifting/points-adding/cluster-adding/object-adding~\cite{xiang2019generating} and point-dropping~\cite{zheng2018learning}) on PointNet. For clean point clouds, DUP-Net slightly reduce 2\% detection accuracy; for C\&W attacks, DUP-Net performs better than baseline SRS, proposed SOR and proposed PU-Net. The DUP-Net performs better than PU-Net, implying that the outlier removing operation is effective in promoting defense performance. For point-dropping attacks, DUP-Net defense is slightly worse than PU-Net alone but still better than baseline, which implies that large modifications made by attackers cause some local shapes to disappear to a large degree, and SOR defense further breaks critical skeleton information. For C\&W cluster and C\&W object attacks, our DUP-Net defense improves the accuracies to 87.6\% and 68.4\%; with SRS defense, the accuracies are 92.0\% and 92.4\%. These results further demonstrate our strong defense ability. Because DUP-Net removes outliers of the manifold surface and SRS equiprobably removes points, for these two tasks, SRS performs better.

Overall, DUP-Net as a preprocessing network help ensure the robustness of neural network based classification and resist attacks from adversarial point clouds. Besides, DUP-Net is non-differentiable which makes attacker difficult to implement secondary adversarial attacks.

\begin{figure*}
\begin{center}
{\centering\includegraphics[width=1\linewidth]{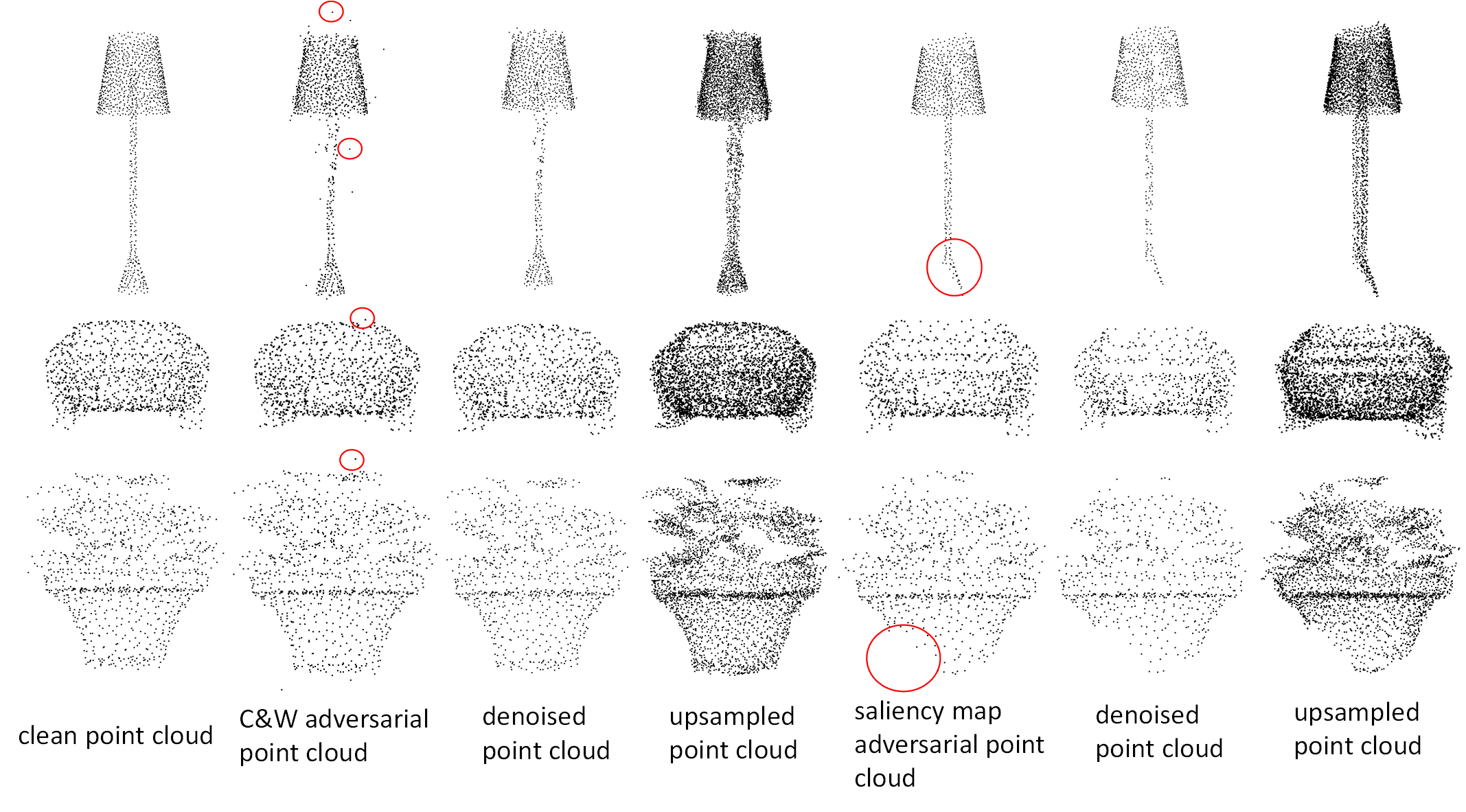}
}\\
\caption{Visualization of point clouds. The fifth column is the 200 points dropping attack. Red circles denote outliers and missing object parts. Enlarge to see details. }
\label{fig:05}
\end{center}
\end{figure*}

\subsection{Generability of DUP-Net}

The transferability of targeted C\&W based points-shifting and points-adding attacks and saliency map based points-dropping attack of PointNet on black-box classification systems is shown in Table~\ref{tab:02}. Similar to \cite{xiang2019generating}, we test the success rate of adversarial examples on PointNet++ and DGCNN. The result illustrates that C\&W based adversarial samples have limited transferability; while for points-dropping attacks, they are more transferable to other networks. We find that for black-box defense, adding a defense network (SRS layer or DUP-Net) before the classification network deteriorates the classification accuracy, thus our proposed defense scheme is only suitable for white-box attack, and in such cases, the defense sides do not need the preprocessing network. The results show that deploying DUP-Net does not improve defense ability against black-box attacks owning to the network structure dissimilarity.

In Table~\ref{tab:03}, we implement the experiments on PointNet++, which further shows our proposed DUP-Net can be utilized for different target recognition models.

\subsection{Visualization}
Figure~\ref{fig:05} shows the details of clean point clouds, adversarial point clouds and the denoised and upsampled point clouds. From top to bottom the classes of point clouds are ``vase'', ``car'' and ``flower pot'', respectively. It shows that C\&W attacks produce outliers that lie far away from the manifold of points, while saliency map based attacks drop a cluster of points. We denote the outliers and dropped cluster by red circles. The SOR denoiser drops some outliers to neutralize the attack success rates of adversarial point clouds, and then PU-Net strengths the smoothness of local region to assist the classification of models.

\section{Conclusion}

In this paper, we propose a denoiser and upsampler network (DUP-Net) formed by a statistical outlier removal (SOR) layer and a point cloud upsampling network (PU-Net) to defend against 3D point cloud adversarial examples. We propose to use point cloud restoration techniques to purify perturbed point clouds. As an initial step, we apply SOR to suppress any outlier based noise patterns and formulate a non-differentiable layer that is difficult to bypass. The central component of our approach is the upsampling operation, which enhances the point resolution while simultaneously removing adversarial patterns similar to image super-resolution operation. Our experiments show that point cloud upsampling network alone is sufficient to reinstate classifier beliefs towards correct categories; besides, the statistical outlier removal step provides added robustness since it is a non-differentiable denoising operation.


\textbf{Acknowledgement} This work was supported in part by the Natural Science Foundation of China under Grant U1636201 and 61572452, and by Anhui Initiative in Quantum Information Technologies under Grant AHY150400.

{\small
\bibliographystyle{ieee}
\bibliography{mybib}
}

\end{document}